# SatAOI: Delimitating Area of Interest for Swing-Arm Troweling Robot for Construction

Jia-Rui Lin[1,2], Shaojie Zhou[1,2], Peng Pan[1,2,*], Ruijia Cai[3] and Gang Chen[4]

[1] Department of Civil Engineering, Tsinghua University, China
[2] Key Laboratory of Digital Construction and Digital Twin, Ministry of Housing and Urban-Rural Development, China
[3] Institute of Architecture Design and Research, Chinese Academy of Sciences, China
[4] China Construction Third Engineering Bureau Beijing Co., Ltd., China

**Abstract**

In concrete troweling for building construction, robots can significantly reduce workload and improve automation level. However, as a primary task of coverage path planning (CPP) for troweling, delimitating area of interest (AOI) in complex scenes is still challenging, especially for swing-arm robots with more complex working modes. Thus, this research proposes an algorithm to delimitate AOI for swing-arm troweling robot (SatAOI algorithm). By analyzing characteristics of the robot and obstacle maps, mathematical models and collision principles are established. On this basis, SatAOI algorithm achieves AOI delimitation by global search and collision detection. Experiments on different obstacle maps indicate that AOI can be effectively delimitated in scenes under different complexity, and the algorithm can fully consider the connectivity of obstacle maps. This research serves as a foundation for CPP algorithm and full process simulation of swing-arm troweling robots.

## 1 Introduction

With the development of automation technology, the application of robots in the construction industry is becoming increasingly widespread [1]. On the one hand, construction robots can replace workers to complete detrimental or monotonic tasks. And on the other hand, robots equipped with high-precision sensors can improve productivity and quality [2], which makes construction robots an important part of achieving intelligent construction. Currently, common on-site construction robots are qualified to perform tasks such as bricklaying, wall painting, building components assembly, and concrete construction [6]. Take concrete floor troweling as an example, to ensure sufficient flatness, floor troweling requires plenty of workers to draw trowels over large surfaces multiple times, which is time-consuming, labour-intensive, and the quality dependent on worker's skills [9]. However, if using a troweling robot, the workload will be greatly reduced, and the surface quality will be easily controlled and consistently ensured [10].

However, the environment at construction sites is often variable and unstructured, posing significant challenges to the automation of construction robots [11]. In order to adapt to different work environments, robots have adopted different designs for high efficiency. Figure 1 shows two different forms of concrete troweling robots. Swing-arm robot is a commonly used concrete troweling robot, and the design of the swing arm enhances the robot's working range while maintaining a compact size. Compared to other types of robots, swing-arm robots exhibit greater flexibility in working modes, enabling them to adapt to a wider range of construction scene. But with an additional degree of freedom for swinging arm, the operation of them becomes more complex. These issues make it difficult to operate troweling robots automatically, and human-robot collaboration [14] or complex pre-defined paths [10] are usually adopted. In addition, the automation of agricultural robotics also faces these issues, but irregular field shapes are often considered as the main constraint of agricultural robots [15]. There usually few obstacles in farmyard but many in construction sites. It will be more challenging when there are many obstacles or holes in the environments. A large number of efforts are needed to improve the automation level of troweling robots.

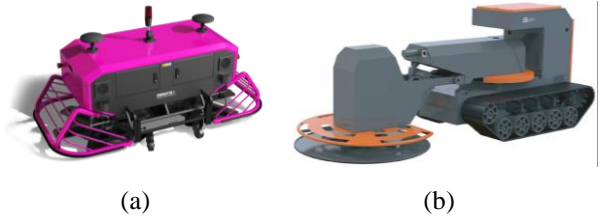

Figure 1. (a) Ground troweling robot of Bright Dream Robotics [12]. (b) Crawler troweling robot of FangShi Technology [13].

The essence of troweling work is the coverage path planning (CPP) problem within a certain scene, that is, to find a collision-free trajectory to ensure the robot fully covers an area of interest (AOI) [16]. The most classical method for CPP is cellular decomposition [16]. Firstly, AOI needs to be effectively delimitated and then decomposed into a series of cells. The final coverage path can be solved through simple planning within the cells and connections between the cells. Therefore, the delimitating AOI is a primary and critical part for this method [17], which will affect all subsequent processes. Furthermore, since the troweling and movement of the swing-arm robot are controlled by different components, the passability and workability between obstacles are complex, which poses a huge challenge to the delimitation of AOI.

In order to improve the troweling efficiency, this research focuses on how to delimitate AOI of swing-arm troweling robot (Figure 1 b). By analyzing the characteristic of robot, **AOI** delimitation algorithm for **S**wing-**A**rm **T**roweling robot (**SatAOI** algorithm, a wish for robots working on Sat.) is constructed for. The remainder of this paper is organized as follows. Section 2 introduces methodology of overall research. Section 3 builds the models of obstacle map and swing-arm robot. Section 4 proposes the SatAOI algorithm. Section 5 presents a case study. Section 6 summarizes the research and discusses possible future investigations.

## 2 Methodology

The objective of this research is to delimitate the AOI of a swing-arm troweling robot. To achieve this objective, it is necessary to fully analyze the working characteristics of the swing-arm robot and establish accurate collision principles. On this basis, AOI is determined based on proposed SatAOI algorithm. The overall methodology is shown in Figure 2.

The first step is to establish the mathematical models of the obstacle map and the troweling robot, of which the most critical part is the definition of collision principles for the swing-arm robot. Then, SatAOI algorithm, composed of collision detection algorithm and search algorithm, is built for AOI delimitation. Finally, a case study is conducted to validate the feasibility of the proposed method. And the AOI results are discussed and visualized for further applying in other downstream tasks such as coverage path planning.

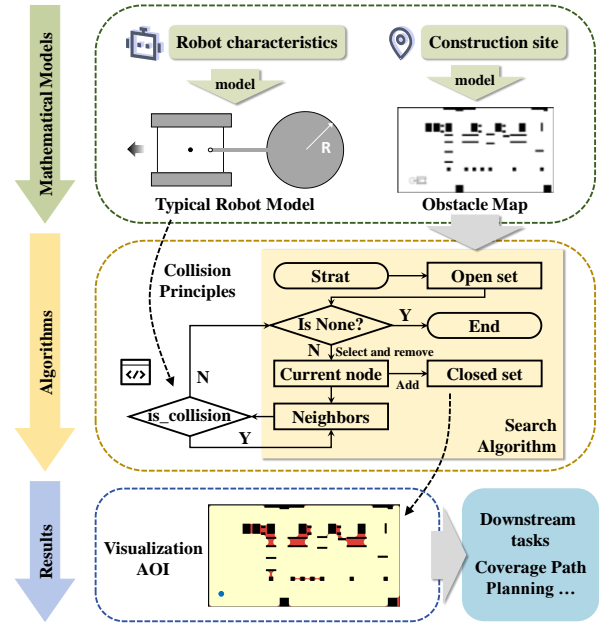

Figure 2. Overall Methodology

## 3 Model Definition

The challenge of delimitation AOI lies in the complexity of obstacle maps and troweling robots. Therefore, it is necessary to first simplify and define the mathematical model of these two research objects to ensure that the algorithm can accurately judge the robot's accessibility and workability according to certain collision principles.

### 3.1 Obstacle Map Model

In concrete troweling for building construction, obstacles refer to indoor building components that have already been constructed (such as walls, columns, etc.) or embedded parts and holes reserved on the ground for building function. This information is usually saved at CAD drawings or BIM models. But these structured data is not conducive for path planning. In continuous 2D environments, obstacle and map information is usually discretized into regular grids (such as equilateral polygons) for robot navigation [18]. These grids are easy to construct and clearly represent blocked and unblocked status. For convenience, squares with a side length of 10 mm are used to redraw the obstacle maps from CAD drawings in our research. As shown in Figure 3, the CAD drawing is converted to a grid-based map. In grid-based map, the cells painted white indicate that areas are accessible. While the cells painted black indicate obstacles, corresponding to areas that robots cannot reach in the real environment, such as walls, columns, pre-embedded parts, holes, etc.

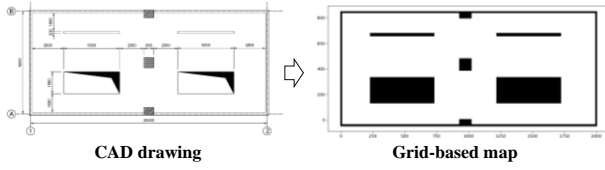

Figure 3. Obstacle map from CAD drawing

## 3.2 Robot Model

In this research, establishing an accurate and comprehensive mathematical model of swing-arm robot is the chief task. As shown in Figure 4, in our research, swing-arm robot is simplified into three parts: robot body (tracked vehicle), swinging arm and troweling disc. The body of robot is a racked vehicle, which controls the movement of robot. Tracked vehicle relies on the differential speed of its tracks for steering. Commonly, the geometric center of vehicle is used as the vehicle center and as well as control point of whole robot. The troweling disc is the main component for working, which can rotate around the swing-arm center within certain angle range. Notably, vehicle center and swing-arm center may not coincide. In order to ensure that the troweled areas are not disturbed, the robot's heading direction is the orientation of the tracked vehicle.

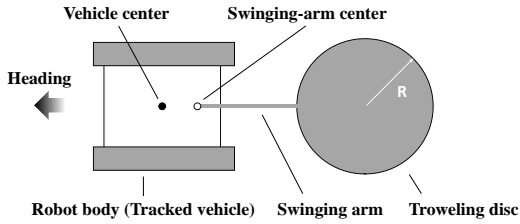

Figure 4. Swinging-arm robot model

As shown in Figure 5, the basic parameters of robot are defined: $V_l$ and $V_w$ represent the length and width of tracked vehicle; $L_{vs}$ represents the offset between vehicle center and swing arm center; $L_s$ represents the length of the swing arm; $R$ represents the radius of troweling disc. Furthermore, if size of robot is determined, accurately defining the position of a swing-arm robot requires 4 parameters: x, y coordinate, yaw, and swing-arm angle $(x, y, yaw, s)$. the x and y coordinates represent the accurate location of vehicle center ($V_c$). The yaw is the angle between robot heading vector and positive x-axis direction, ranging from 0 to $2\pi$ (counterclockwise is positive). The swing-arm angle represents the angle deviation of the swing arm from its original position (counterclockwise is positive), and the range depends on the design parameters of robots. For convenience, this angle is specified as $[-\pi/2, \pi/2]$. The above four parameters are sufficient to definite any position of the robot. In a certain position, the effective work area is the coverage area of troweling disc. For convenience, use $S_c: (S_x, S_y)$ and $D_c: (D_x, D_y)$ to represent the center of swinging arm and troweling disc, separately. Then the relationship between $V_c$, $S_c$ and $D_c$ can be expressed as follows:

$$V_c = S_c + L_{vs} \cdot h \quad (1)$$
$$S_c = D_c + L_s \cdot (\cos(yaw + s), \sin(yaw + s)) \quad (2)$$

in which $h$ represent the unit vector of heading direction:

$$h = (\cos(yaw), \sin(yaw)) \quad (3)$$

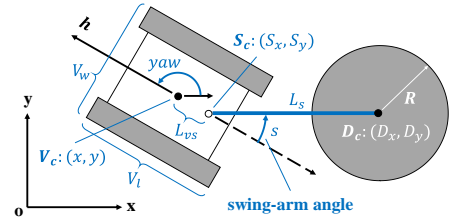

Figure 5. Four parameters for robot positioning: x, y, yaw, and swing-arm angle

On this basis, the collision principles of swing-arm robot are defined. Given that the tracked vehicle part and swinging arm part have different geometric shape, two kinds of collision principles are employed separately. As shown in Figure 6, if the position of an obstacle point is denoted as $O_c: (O_x, O_y)$. In terms of tracked vehicle, collision detection is divided into two rounds. In the first round, the vehicle's circumscribed circle is used to exclude all obstacles outside the circle (obstacle No.1 in Figure 6, Equation 4).

$$\|O_c - V_c\|_2 \geq \frac{\sqrt{V_w^2 + V_l^2}}{2} \quad (4)$$

If any obstacles are inside the circle, proceed to the second round. By calculating the projections of the vector from the vehicle's center to the obstacle in both the heading direction and the perpendicular direction, the collision status can be determined. If the absolute values of these two projections are simultaneously less than half the vehicle's length and width, respectively, the obstacle is judged within the collision range (obstacle No.3 in Figure 6, Equation 5&6). Otherwise, it will not result in a collision (obstacle No.2 in Figure 6).

$$|(O_c - V_c) \cdot h| \leq V_l/2 \quad (5)$$

$$\sqrt{\|O_c - V_c\|^2 - |(O_c - V_c) \cdot h|^2} \leq V_w/2 \quad (6)$$

As for swinging arm and disc, the collision cannot be judged by a simple circle or square. As shown in Figure 6, the avoidance area of this part is a capsule-like shape. This area can be solved by calculating the shortest distance from the obstacles to the line segment **Sc**-**Dc**. This process is also divided into 2 rounds. Firstly, eliminate most irrelevant obstacles through a circle externally connected to the avoidance area (obstacle No.4 in Figure 6, Equation 7).

$$\|O_c - C\|_2 \geq \frac{L_s + R}{2} \quad (7)$$

in where $C$ is the center of avoidance area. As for the remaining obstacles, if they are within the perpendicular range of the endpoints of the line segment $D'_c$-$D_c$ ($D'_c$ is the $D_c$ about the symmetry point of $C$), such as obstacle No.5 in Figure 6 (Equation 8).

$$|(O_c - C) \cdot h'| \leq \frac{L_s - R}{2} \quad (8)$$

in where $h'$ is the unit vector of $S_c$ to $D_c$.

$$h' = \frac{D_c - S_c}{|D_c - S_c|} \quad (9)$$

In this case, their perpendicular distance to the line $D'_c$-$D_c$ needs to be greater than $R$ to ensure no collision (Equation 10).

$$\sqrt{\|O_c - C\|^2 - |(O_c - C) \cdot h'|^2} > R \quad (10)$$

If obstacles don't satisfy the Equation 8, such as No.6&7 in Figure 6. No collision means that the minimum distance between obstacle and $D'_c/D_c$ is greater than $R$ (Equation 11).

$$\min(\|O_c - D'_c\|_2, \|O_c - D_c\|_2) > R \quad (11)$$

Notably, in the actual algorithm, all calculations can take redundancy into account to ensure sufficient safety.

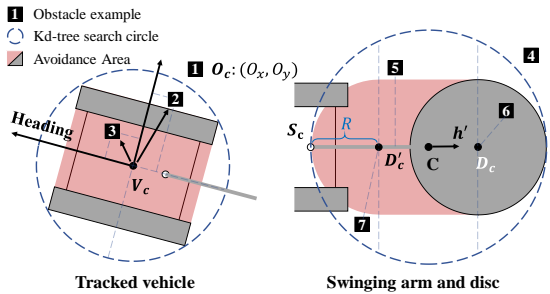

Figure 6. Collision principles of tracked vehicle, swinging arm and disc

## 4 SatAOI Algorithm for Delimiting AOI

In complex environments with a large number of pre-embedded parts and holes, delimitation of AOI will become more difficult and important. At the same time, since the working range of the troweling disc is separate from the main body of the robot, it further increases the challenges in the delimitation of AOI for the swing-arm robot. So, in this research, using the troweling disc as the main reference, whole map is searched to determine the AOI. The SatAOI algorithm is composed of two core parts: collision detection centered around disc (Algorithm 1: is_collision) and global AOI search (Algorithm 2: troweling_search).

**Algorithm 1: is_collision.** As shown in Figure 7, if troweling disc is put at certain coordinate, while swing-arm and vehicle are rotated to $angle1$ and $angle2$, separately, the $yaw$ and swinging angle ($s$) can be calculated by Equation 12&13.

$$yaw = angle1 + angle2 \quad (12)$$
$$s = -angle2 \quad (13)$$

Then based on the robot model and collision principles in 3.2, the collision can be inferred by algorithm 1. If collision occurs, "True" is output. Otherwise, "False" is output.

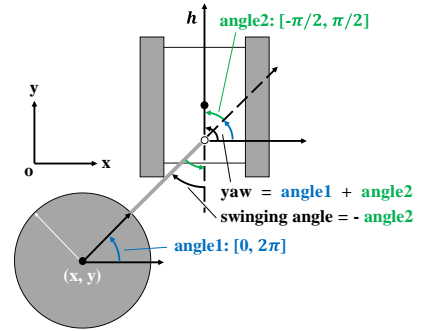

Figure 7. Collision detection algorithm using the troweling disc as the main reference

| **Algorithm 1**: is_collision($D_c$, $angle1$, $angle2$) |
|---|
| **Input**: Coordinate of disc center $D_c$: $(x, y)$, rotate angles of swing-arm and vehicle $angle1, angle2$. |
| **Output**: If there is a collision, return True, else False. |
| **1**  // kdtree_o is specifically for storing obstacles |
| **2**  $kdtree\_o \leftarrow$ kd.KDTree(obstacle grids) |
| **3**  **If** $kdtree\_o$.query_ball_point($D_c$,$R$) **then** |
| **4**     **Return** True |
| **5**  $C \leftarrow$ center of capsule-like shape in Figure 6 |
| **6**  // $R$ is radius of disc |
| **7**  $obstacle1 \leftarrow kdtree\_o$.query_ball_point($C, \frac{L_s+R}{2}$) |
| **8**  **If** $obstacle1$ **then** |

| | |
|---|---|
| 9 | $h'$ ← unit vector of $S_c$ to $D_c$. (Equation 9) |
| 10 | **For** $p$ in *obstacle1* **do** |
| 11 |   $vec$ ← $p - C$ |
| 12 |   **If** $\lvert vec \cdot h' \rvert \leq \frac{L_s - R}{2}$ **then** |
| 13 |     **If** $\sqrt{\lVert vec \rVert^2 - \lvert vec \cdot h' \rvert^2} \leq R$ **then** |
| 14 |       **Return** True |
| 15 |   **Else** |
| 16 |     **If** $\min(\lVert p - D_c' \rVert_2, \lVert p - D_c \rVert_2) \leq R$ **then** |
| 17 |       **Return** True |
| 18 | $V_c$ ← Center of tracked vehicle |
| 19 | $r$ ← $\sqrt{V_w^2 + V_l^2}$ / 2 |
| 20 | *obstacle2* ← *kdtree_o*.query_ball_point($V_c$, $r$) |
| 21 | **If not** *obstacle2* **then** |
| 22 |   **Return** False |
| 23 | **Else** |
| 24 |   **For** $p$ in *obstacle2* **do** |
| 25 |     $vec$ ← $p - V_c$ |
| 26 |     $yaw$ ← $angle1 + angle2$ |
| 27 |     $h$ ← $(\cos(yaw), \sin(yaw))$ |
| 28 |     $proj$ ← $vec \cdot h$ |
| 29 |     $proj\_v$ ← $\lVert vec \rVert^2 - proj^2$ |
| 30 |     **If** $proj < V_l/2$ **and** $proj\_v < V_w/2$ **then** |
| 31 |       **Continue** |
| 32 |   **Return** False |

**Algorithm 2: troweling_search.** Inspired by BFS (Breadth-first search) algorithm [19], the search process is a step-by-step exploring that spreads from one point to the surrounding areas. There are two sets: open set is points need to explore, and closed set is points already explored. Starting from a certain point (start node), the center of troweling disc is put at this point. Firstly, add the start node to the open set. Then, loop begins. While the open set is not empty, select the arbitrary node in the open set as the current node (also move it from the open set to the closed set). For each neighbor of the current node (other eight nodes centered around the current node in nine-square grid), if the neighbor is in the closed set, skip it as it has already been evaluated. Otherwise, stepping by $\pi/12$ and considering all discrete angle combinations ($24 \times 13 = 312$ pairs) of angle1&2 at this node, if there a at least a pair of certain angles that frees robot from collision, add this node to open set. Finally, all points that are connected to the start point and can be used to place a troweling disc (robot) are left in the closed set. So, this search algorithm can consider the connectivity of obstacle map.

| | |
|---|---|
| **Algorithm 2**: troweling_search(*start_node*) | |
| **Input**: The starting point of search *start_node*. | |
| **Output**: area of interest *AOI* | |
| 1 | // kdtree_v is specifically for storing points excluding obstacles |
| 2 | *kdtree_v* ← kd.KDTree(all grids excluding obstacle grids) |
| 3 | // Points need to explore *open_set*, points already explored *closed_set* |
| 4 | *open_set* ← {*start_node*}, *closed_set* ← {} |
| 5 | **While** *open_set* is not empty **do** |
| 6 |   *current_node* ← Arbitrary node in *open_set* |
| 7 |   Remove *current_node* from *open_set* |
| 8 |   **For** *node* in neighbors of *current_node*: |
| 9 |     **If** *node* in *closed_set* **then** |
| 10 |       **Continue** |
| 11 |     *areso* ← $\pi/12$ // angular resolution |
| 12 |     *angle1* ← range(0, $2\pi$, *areso*) |
| 13 |     *angle2* ← range(-$\pi/2$, $\pi/2$, areso) |
| 14 |     **For** *a1*, *a2* ← list(zip(*angle1*, *angle2*)) **do** |
| 15 |       **If** is_collision(*node*, *a1*, *a2*) **then** |
| 16 |         **Continue** |
| 17 |       **Else** |
| 18 |         Add *node* to *open_set* |
| 19 |         **Break** |
| 20 | **For** *node* in *closed_set* **do** |
| 21 |   **For** *point* in *kdtree_v*.query_ball_point(*node*, *cR*) **do** |
| 22 |     Add *point* to *AOI* |
| 23 | **Return** *AOI* |

Based on SatAOI algorithm, all possible positions of the troweling disc have been determined, and the envelope region composed of all discs is the final theoretical limit AOI.

## 5 Case Study

In order to verify the feasibility of the proposed method, taking architectural floor plans of a substation under different complexity as research cases, the robustness of SatAOI algorithm for different scenes and its effectiveness for connected and non-connected areas are tested. Based on this, the importance of AOI delimitation for coverage path planning is discussed.

### 5.1 Troweling Robot Parameters

The troweling robot from MagicBIM is selected as the robot prototype for simulation. As shown in Figure 8, the tracked vehicle is simplified into a rectangle with a length ($V_l$) of 875mm and a width ($V_w$) of 830mm. The center of swing arm deviates from the center of vehicle ($L_{vs}$) by 142mm, the length of the swing arm ($L_s$) is 995mm, and the radius of the troweling disc ($R$) is 350mm. At the same time, the robot can achieve stationary turning by rotating the tracks on both sides in opposite directions. And the maximum swing-arm range is $[-\pi/2, \pi/2]$. These parameters are sufficient to characterize a typical swing-arm troweling robot, including the basic rules for motion, collision and troweling.

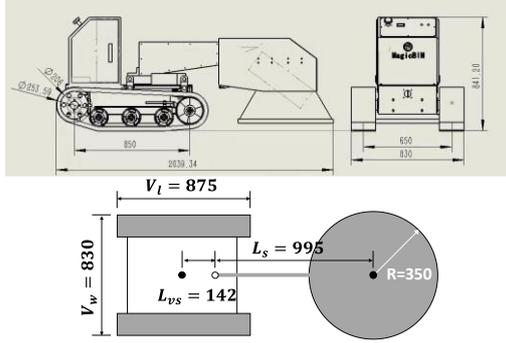

Figure 8. Prototype and key parameters of robot

### 5.2 Obstacle Maps Generation

The gas insulated switchgear (GIS) room is an important part in substations. However, a large number of electrical equipment requires a large number of embedded parts and holes to be reserved during the construction process, which poses a huge challenge for automation of troweling robots. So, using CAD drawings of different GIS rooms as the data source, obstacle maps under different complexity are generated for testing. As shown in Figure 9, floor plans are converted into grid-based maps using squares with a side length of 10mm. And there are three complexity levels of obstacle maps, denoted as "simple map", "medium map" and "complex map", separately. The overall size of three scenes is same of 10,450mm * 19,190mm. And size of Obstacles (embedded parts and holes) varies from 160mm to 2000mm. Notably, in order to test the ability of SatAOI algorithm to recognize map connectivity, a bottom-to-top block is added to the "simple map" and divides the map into left and right parts. According to statistics, the area proportions of obstacle in each map are shown in Table 1. The area of entire map reached 200.54 m$^2$, and the highest obstacle area proportion is 12.89%.

Table 1. Statistics of obstacles

| Map type | Map area (m$^2$) | Obstacle area (m$^2$) | Proportion (%) |
|---|---|---|---|
| Simple |  | 7.00 | 3.49 |
| Medium | 200.54 | 14.96 | 7.46 |
| Complex |  | 25.84 | 12.89 |

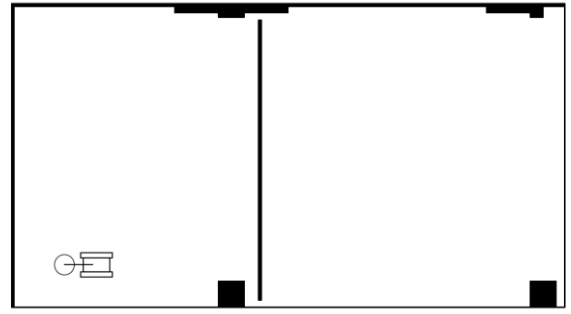

(a) Simple map with a block

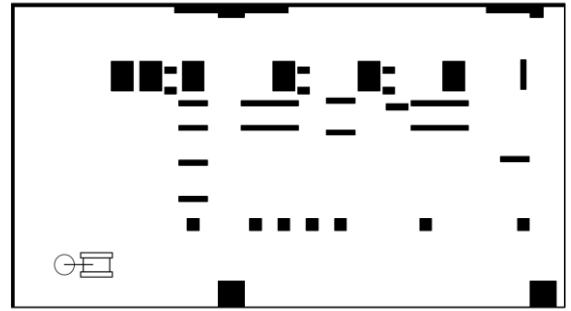

(b) Medium map

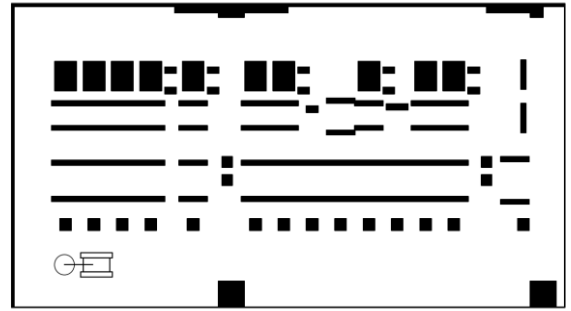

(c) Complex map

Figure 9. Obstacle maps and robots. (a) (b) (c) represent maps under different complexity

In order to demonstrate challenges of obstacles to the swing-arm troweling robot in a more intuitive way, robots and obstacle maps are simultaneously visualized in Figure 9. It is evident that the complex obstacles in GIS room pose significant challenges to the swing-arm troweling robots, even if manual operations are also extremely difficult in this case. There is no doubt that the delimitation of AOI can assist in the planning and simulation of swing-arm troweling robots.

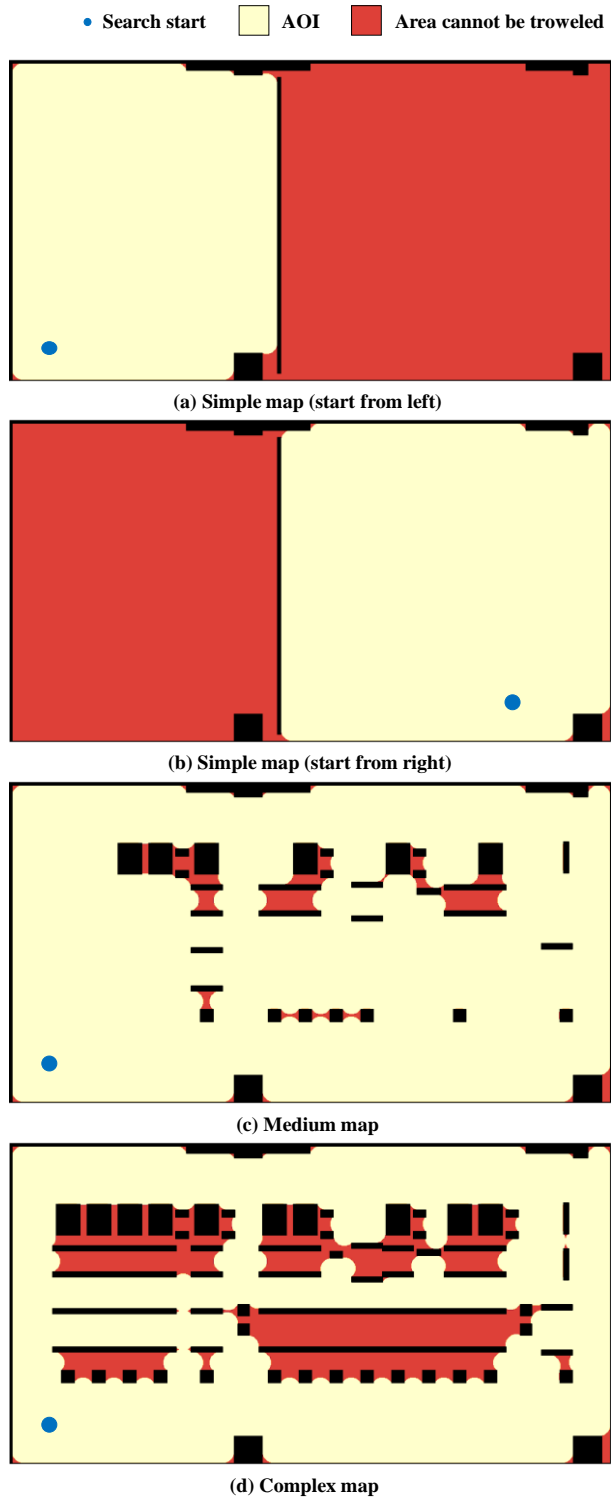

(a) Simple map (start from left)

(b) Simple map (start from right)

(c) Medium map

(d) Complex map

Figure 10. AOI results of different obstacle maps

### 5.3 AOI Delimitation

The results of SatAOI algorithm are shown in the Figure 10, in which the yellow areas represent the final AOI in all maps. Overall, the all results of AOI are reasonable and effective, only narrow gaps between obstacles and wall corners cannot be troweled by robots. Regardless of the complexity of maps, AOI results effectively represent the maximum troweling areas of the swing-arm robot. In complex maps (Figure 10 d), the algorithm can identify which areas between obstacles can be troweled and which cannot, bringing great convenience to robot simulation. Furthermore, in the simple map (Figure 10 a&b), AOI include left part and right part due to different search starts. This is because the left and right sides of the map are separated by a bottom-to-top block. And in reality, the robot cannot complete the troweling work of both parts at once. The SatAOI algorithm can identify the connectivity between different areas, ensuring that the AOI result can be troweling by the robot in a single task.

These analysis results can be applied to some downstream tasks for construction robots, such as assisting in the path planning and simulation of swing-arm troweling robot in complex construction sites. According to the AOI results, the robot can determine which areas can be reached and troweled, and which areas will not be accessed. On this basis, AOI can be further divided into small cells for CPP task. AOI is the goal of coverage path optimization, and will also participate in the evaluation process of path planning.

## 6 Conclusion

Although robots have improved the mechanization and automation of construction, swing-arm troweling robot still rely on manual operation in complex construction sites. In this research, the mathematical models of swing-arm robot and obstacle map are built. On this basis, collision detection principle and search method are proposed, together formed the SatAOI algorithm for the AOI delimitation of swing-arm troweling robot. A case study of GIS room from a real substation construction is conducted and the results show that SatAOI algorithm can effectively determine the AOI in maps under different complexity, and meanwhile can distinguish the connected and disconnected areas from a certain start.

The delimitation of AOI is only a first and primary task of troweling robot simulation. Further research is the downstream application of AOI. Cellular decomposition method or other CPP algorithm based on AOI results will greatly enhance the automation level of swing-arm troweling robots. AOI-based coverage path optimization and evaluation can further improve the work efficiency of the troweling robot. In addition, even if the robot cannot be fully automated, AOI can assist workers in achieving more efficient human-robot collaboration and improving construction productivity [20].

# 7 Acknowledgement

This research was funded by the National Key R&D Program of China (grant No. 2023YFC3805800), the project "Research on Key Technologies for Mechanized Construction of Substation Civil Engineering" of "Technology project funding from State Grid Corporation of China" (project No. 5200-202311481A-3-2-ZN) and the project "Constructing a theoretical system for sustainable development strategy of construction enterprises with digital platforms as the core" from China Construction Third Engineering Bureau Beijing Co., Ltd.